\documentclass[acmtog]{acmart}
\acmSubmissionID{1032}
\usepackage{wrapfig}
\usepackage{xcolor}
\usepackage{subcaption}

\usepackage[ruled]{algorithm2e} 

\SetAlFnt{\small}
\SetAlCapFnt{\small}
\SetAlCapNameFnt{\small}
\SetAlCapHSkip{0pt}

\acmJournal{TOG}

\copyrightyear{2024}
\acmYear{2024}
\setcopyright{acmlicensed}\acmConference[SA Conference Papers '24]{SIGGRAPH Asia 2024 Conference Papers}{December 3--6, 2024}{Tokyo, Japan}
\acmBooktitle{SIGGRAPH Asia 2024 Conference Papers (SA Conference Papers '24), December 3--6, 2024, Tokyo, Japan}
\acmDOI{10.1145/3680528.3687682}
\acmISBN{979-8-4007-1131-2/24/12}

\setcitestyle{square}

\usepackage{dsfont}

\def\dd{\mathbf{d}}

\def\ii{\mathbf{i}}

\def\mm{\mathbf{m}}
\def\nn{\mathbf{n}}

\def\pp{\mathbf{p}}
\def\qq{\mathbf{q}}

\def\xx{\mathbf{x}}
\def\yy{\mathbf{y}}

\def\dD{\mathcal{D}}

\def\mM{\mathcal{M}}
\def\nN{\mathcal{N}}

\def\sS{\mathcal{S}}

\def\xX{\mathcal{X}}

\def\Re{\mathbb{R}}

\DeclareMathOperator*{\argmax}{arg\,max}

\def\eps{\varepsilon}

\newcommand\paren[1]{\left(#1\right)}

\newcommand\norm[1]{\left\lVert#1\right\rVert}

\DeclareMathSymbol{@}{\mathord}{letters}{"3B}

\def\latex/{\LaTeX}
\def\bibtex/{\hologo{BibTeX}}

\newcommand{\RN}[1]{%
  \textup{\uppercase\expandafter{\romannumeral#1}}%
}

\usepackage{multirow}

\makeatletter
\DeclareRobustCommand\onedot{\futurelet\@let@token\@onedot}
\def\@onedot{\ifx\@let@token.\else.\null\fi\xspace}
\def\eg{\emph{e.g}\onedot} 
\def\ie{\emph{i.e}\onedot}

\def\etal{\emph{et al}\onedot}
\makeatother

\usepackage[capitalize]{cleveref}
\Crefname{section}{Section}{Sections}
\Crefname{table}{Table}{Tables}
\Crefname{tab}{Table}{Tables}
\Crefname{figure}{Figure}{Figures}
\Crefname{fig}{Figure}{Figures}

\newcommand{\proj}[0]{\operatorname{proj}}
\usepackage{hyperref}

\begin{document}

\title{Robust Symmetry Detection via Riemannian Langevin Dynamics}

\author{Jihyeon Je}
\authornote{Equal contribution}
\email{jihyeonj@stanford.edu}
\affiliation{%
 \institution{Stanford University}
 \city{Stanford}
 \country{USA}}
\author{Jiayi Liu}
\authornotemark[1]
\email{jiayiliu@stanford.edu}
\affiliation{%
 \institution{Stanford University}
 \city{Stanford}
 \country{USA}}
\author{Guandao Yang}
\authornotemark[1]
\email{guandao@stanford.edu}
\affiliation{%
 \institution{Stanford University}
 \city{Stanford}
 \country{USA}}
\author{Boyang Deng}
\authornotemark[1]
\email{bydeng@stanford.edu}
\affiliation{%
 \institution{Stanford University}
 \city{Stanford}
 \country{USA}}
\author{Shengqu Cai}
\email{shengqu@stanford.edu}
\affiliation{%
 \institution{Stanford University}
 \city{Stanford}
 \country{USA}}
\author{Gordon Wetzstein}
\email{gordonwz@stanford.edu}
\affiliation{%
 \institution{Stanford University}
 \city{Stanford}
 \country{USA}}
\author{Or Litany}
\email{orlitany@gmail.com}
\affiliation{%
 \institution{Technion}
 \city{Haifa}
 \country{Israel}}
\author{Leonidas Guibas}
\email{guibas@cs.stanford.edu}
\affiliation{%
 \institution{Stanford University}
 \city{Stanford}
 \country{USA}}

%
%
%
%


\begin{teaserfigure}
  \vspace{-1em}
 \includegraphics[width=\textwidth, height=250pt]{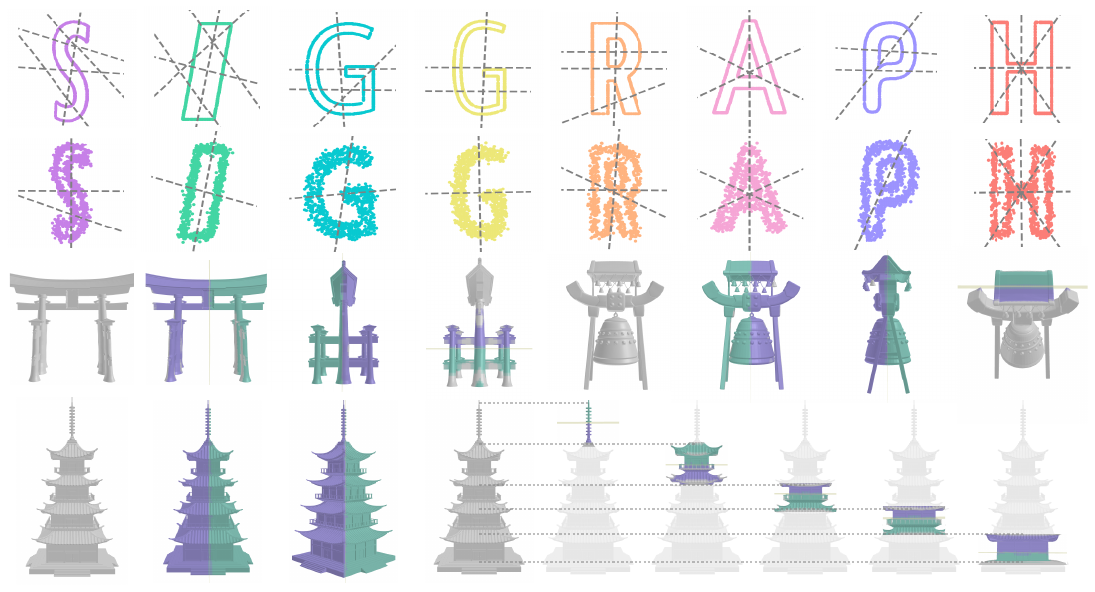}
  \vspace{-2.5em}
  \caption{
  In this work, we present a novel Langevin dynamics-based symmetry detection method. Our solution is robust to noise, and can detect both partial and global reflective symmetries on diverse shapes. Here we visualize detected reflective symmetries at different noise levels for 2D (top half), and illustrate detected planes along with its supports for selected 3D shapes (bottom half).
  }
\label{fig:teaser}
\end{teaserfigure}
\begin{abstract}
Symmetries are ubiquitous across all kinds of objects, whether in nature or in man-made creations. While these symmetries may seem intuitive to the human eye, detecting them with a machine is nontrivial due to the vast search space. Classical geometry-based methods work by aggregating "votes" for each symmetry but struggle with noise. In contrast, learning-based methods may be more robust to noise, but often overlook partial symmetries due to the scarcity of annotated data. In this work, we address this challenge by proposing a novel symmetry detection method that marries classical symmetry detection techniques with recent advances in generative modeling. Specifically, we apply Langevin dynamics to a redefined symmetry space to enhance robustness against noise. We provide empirical results on a variety of shapes that suggest our method is not only robust to noise, but can also identify both partial and global symmetries. Moreover, we demonstrate the utility of our detected symmetries in various downstream tasks, such as compression and symmetrization of noisy shapes.{\renewcommand\UrlFont{\color{magenta}\rmfamily\itshape} Please see our project page and code at \url{https://symmetry-langevin.github.io/}.}

\end{abstract}


\maketitle
\section{Introduction}
\label{sec:intro}
Symmetry, being an aesthetically beautiful concept per se, is fundamental to our understanding of structures in geometry. According to the great mathematician Blaise Pascal, it is essentially ``what we see at a glance''.
Therefore, the problem of discovering symmetries in shapes has been a central pursuit in geometry processing~\cite{atallah1985symmetry, wolter1985optimal, mitra2006partial, pauly2008discovering}.
The capability of automatically detecting symmetries can consequently enable us to compress~\cite{pauly2008discovering} and symmetrize~\cite{mitra2013symmetry} shapes for numerous uses in shape modeling, manufacturing, or artistic creation.

Albeit an intuitive concept, symmetry is not trivial to detect, especially when both global and partial symmetries are considered.
The challenge of symmetry detection lies in the vast search space of this problem -- one has to determine not only the symmetries present but also their support in the shape geometry.
The brute force solution has been to search over pairs of points on the shape surface and ascertain whether their neighbouring regions may indicate the presence of a symmetry.
To address this complexity, \citet{mitra2006partial} propose to map the problem from the shape space to a transformation space, where each point is the transformation between a point pair sampled on the original shape surface.
In this transformation space, symmetries can be found by identifying dense clusters, which they do by mean-shift clustering. 
However, while this largely reduces the search space, it is built on the assumption that the original shape is noise-free.
In contrast, in the real world, more often we face scenarios when we need to process noisy or even incomplete shapes, sourced from imperfect sensors.
Hence, limited robustness to noisy shapes can significantly restrict the applicability of a symmetry detection algorithm.
On the other hand, attempts have been made to design symmetry detection algorithms by learning from data ~\cite{ji2019symnet, gao2020prs, zhou2021nerd}. 
While this improves robustness to data noise, the lack of annotated data, particularly for partial symmetries, unfortunately limits these methods to detecting only global symmetries.

In this work, we aim to design a symmetry detection algorithm that is robust to noise, while being capable of identifying symmetries of different strengths. We extend the formulation of symmetry detection in the transformation space in \citet{mitra2006partial} and provide a fresh perspective of understanding this transformation space.
Inspired by the recent success of score-based sampling ~\cite{song2021scorebased, karras2022edm, song2020ddim, poole2022dreamfusion} in generative models, we view the transformation space as a score field where the symmetries we seek are the modes in this field.
As robust mode seeking algorithms are well-studied in generative modeling, we introduce one, specifically Langevin dynamics, to symmetry detection.
We hypothesize that such formulation of the transformation space reduces sensitivity to noise, due to smoothing over sampled points. Building on these insights, we propose a robust symmetry detection algorithm that marries elegant ideas in geometry processing and generative models, leveraging Langevin dynamics as the key driver. We show empirical results indicating that our proposed symmetry detection algorithm is more robust to noise compared to other baselines. In addition, our method is capable of detecting both global and partial symmetries, and is easily generalizable to a wide range of 2D and 3D shapes.
\section{Related Works}
\label{sec:related}
\paragraph{Symmetry}
Symmetries play a crucial part in Computer Vision and Computer Graphics.
They are widely used in various tasks including
3D reconstruction~\cite{gordon1990shapefromsymmetry, mukherjee1995shapefromsymmetry, fawcett1994symmetry, huynh1999symmetry, franois2003mirrorss, thrun2005shapefs, sinha2012symmetry, phillips2016seeinggf, chen2018autosweep, tulsiani2020object, xu2020ladybird},
inverse rendering~\cite{wu2020unsupervised},
shape refinement~\cite{mitra2007symmetrization},
and image manipulation~\cite{zhang2020portrait}.
Consequently, symmetry detection has long been an essential challenge in geometry processing. In this regard, we categorize prior methods into two groups, depending on whether data-driven techniques are used.
For the non-learning based group, early works focused on finding exact symmetries on planar point sets~\cite{atallah1985symmetry,wolter1985optimal}. Later on, the seminal work by Mitra \etal~\shortcite{mitra2006partial} reformulated this detection problem as clustering in the transformation space, offering a new perspective that allows the detection of partial and approximate symmetries. Follow-up works further refined this framework to detect regular and repeated geometric structures~\cite{pauly2008discovering} and orbits~\cite{shi2016symmetry}. In contrast, following the revolutionary breakthrough of deep learning, more recent symmetry detection approaches have focused on learning symmetries from data. SymNet~\cite{ji2019symnet} devised a supervised training framework to classify whether points are located on the symmetry plane. Alternatively, SymmetryNet~\cite{shi2020symmetrynet} predicted symmetrical correspondences to avoid overfitting. NeRD~\cite{zhou2021nerd} further introduced a geometry reconstruction phase to verify predicted symmetries at inference time. However, in these supervised frameworks, annotating symmetries has been an expensive bottleneck.
To address this issue, PRSNet~\cite{gao2020prs} proposed an unsupervised symmetry prediction framework using a novel symmetry distance loss,
while E3Sym~\cite{li2023e3sym} developed an unsupervised correspondence prediction framework.
Building upon the aforementioned approaches, our proposed algorithm bridges classical geometry processing techniques with tools from learning-based approaches by interpreting the transformation space as a score field. Using Langevin dynamics, we achieve training-free symmetry detection that is more robust to noisy inputs than heuristic methods, while also enabling partial symmetry prediction—a challenge yet to be addressed by learning-based approaches.

\paragraph{Score-based Generative Models}
Score-based generative models have emerged as a powerful framework for estimating and sampling from unknown data distributions, achieving unparalleled success in diverse domains such as image generation~\cite{ho2020ddpm, song2020ddim, saharia2022imagen, rombach2022latentdiffusion, dhariwal2021adm, karras2022edm}, 3D generation~\cite{poole2022dreamfusion} and video synthesis~\cite{blattmann2023svd, he2022lvdm, ho2022imagenvideo, blattmann2023alignyourlatents, ge2023pyoco, guo2023animatediff, brooks2024sora}.
Initially inspired by non-equilibrium thermodynamics, \citet{sohldickstein2015noneqthermodynamics} proposed learning a generative model by reversing a gradual Gaussian diffusion process.
One of the seminal contributions to score-based generative models is the denoising diffusion probabilistic model (DDPM)~\cite{ho2020ddpm}, which employs two Markov chains: a forward chain that corrupts original data into noise for training purposes, and a reverse chain that restores this noise to data sample during inference. The forward chain is typically designed to methodically convert any data distribution into a simpler prior distribution, commonly a standard Gaussian distribution. The reverse chain undoes this diffusion process by utilizing deep neural networks to learn the necessary transitions. To generate new data points, one begins with a random noise vector and uses ancestral sampling through the reverse Markov chain. Leveraging the DDPM framework, diffusion models have demonstrated impressive capabilities, particularly in image synthesis~\cite{dhariwal2021adm}.
Score-based generative models \cite{song2019estimating, song2019improvedsgm} offer an alternative formulation. Similar to DDPMs~\cite{ho2020ddpm}, these models perturb data with progressively increasing Gaussian noise and estimate the score functions across all noisy distributions. A deep neural network, conditioned on varying noise levels, is trained to estimate the score functions. During inference, samples are drawn by sequentially applying score functions at decreasing noise levels.
DDPMs and score-based models can also be generalized to infinitely many time steps or noise levels, where the perturbation and denoising are modeled as solutions to SDEs. Score-SDE~\cite{song2021scoresde} represents a significant advance by employing a diffusion process described by an SDE to perturb data with noise. The reverse process, modeled by a corresponding reverse-time SDE, gradually transforms noise back into data. With the score function parameterized by a deep neural network, samples can be drawn through various numerical techniques, such as Langevin dynamics~\cite{song2019estimating}, SDE solvers~\cite{song2021scoresde}, and ODE solvers~\cite{song2021scoresde, karras2022edm, lu2022dpm, song2020ddim}.
Subsequent works focus on improving the accuracy and efficiency of score-based generative models. An early work on accelerating diffusion model sampling, Denoising Diffusion Implicit Models~(DDIM)~\cite{song2020ddim} extends the original DDPM~\cite{ho2020ddpm} formulation, transitioning the sampling process to a fully deterministic reversion of this non-Markov perturbation process. Further, EDM~\cite{karras2022edm} demonstrates that Heun's second-order method~\cite{ascher1998ode} offers an balance between sample quality and sampling speed. Its proposed higher-order solver minimizes discretization errors at the expense of an additional evaluation of the learned score function per timestep and shows that using Heun's method~\cite{ascher1998ode} can produce samples with quality on par with or better than Euler's method within fewer sampling steps.

\section{Background}
\label{sec:prelim}

In this section, we introduce the notations and provide the necessary background in symmetry detection to discuss our method.

\subsection{Geometric Symmetry}

Symmetries, one of the most fundamental shape descriptors, can be categorized as either extrinsic or intrinsic depending on which space the symmetry exists in, and global and partial depending on the extent of the region it impacts. 
Despite the ubiquity, symmetries are very challenging to define \cite{shi2016symmetry, mitra2006partial, tevshaung}.
In this paper, we adapt the geometric view to define symmetries as a transformation under which a significant region of the shape is invariant.
Formally, given a shape $S\in\sS$ represented as a set of $\Re^d$ points, symmetry is a transformation $T:\Re^d\to\Re^d$ that can be applied to a shape region $R \subseteq S$:
\begin{align}
    \exists R \subset S, \text{ s.t. } \dD(T(R), R) \leq \eps, 
    \text{ and }M(R) > \tau
\end{align}
where $\dD:\sS\times\sS\to\Re$ is a distance function between two patches, $T(R)$ denotes the transformed point region $T(R) = \{T(p) : p \in R\}$, $M:\sS \to \Re$ measures the significance of the shape patch. 
To make this definition specific, one needs to define the distance function $\dD$ between regions, the class of transformations $T$, the measure of significance $M$, as well as the thresholds $\tau$ and $\eps$. We illustrate the exact formulations below.

An instance of the distance function is Chamfer distance, which leverages the Euclidean distance between points:
\begin{align}
\label{eq:chamfer_distance}
    \dD_{CD} (S_1, S_2) = 
    \frac{1}{|S_1|}\sum_{\xx\in S_1} \min_{\yy\in S_2} \norm{\xx-\yy} + \frac{1}{|S_2|}\sum_{\yy\in S_2} \min_{\xx\in S_1} \norm{\xx-\yy}
\end{align}
Alternatively, one can use shape properties other than extrinsic shape features to compute the distance. For example, intrinsic shape features such as the Laplace–Beltrami eigenfunctions can give rise to the definition of intrinsic symmetries~\cite{ovsjanikov2008global}. One can also use semantic information or geodesic distance to define $\dD$.
For simplicity, we use Chamfer distance as our distance function.

Note that if the transformation class is expressive enough or the tolerance $\eps$ is very large, then any arbitrary pair of points can give rise to a symmetry.
However, it is often the prominent symmetries, such as the global symmetries that can be applied to the entire shape, that are useful in downstream applications. To ensure the detection of prominent symmetries, we define a measure $M$ for patch significance. One common way to define this measure is the area ratio of the patch region relative to the entire shape, $M(R) = |R|/|S|$.
With an adequate threshold $\tau$, we prioritize symmetries that can be applied to a significant enough region of the shape.
Finally, we need to decide on the class of transformation $T$.
In this paper, we focus on perhaps the simplest yet the most common type of symmetry: reflective symmetry, while we also show that the pipeline could be extended to other symmetry classes, such as rotation and translation (\cref{sec:applications}).

\subsection{Detection of Reflective Symmetries}


\setlength{\intextsep}{0pt}
\setlength{\columnsep}{10pt}
\begin{wrapfigure}{r}{0.27\textwidth}
\centering
\vspace{-1pt}
\includegraphics[width=\linewidth, trim={0.95cm 0cm 0.0cm 0.85cm}, clip]{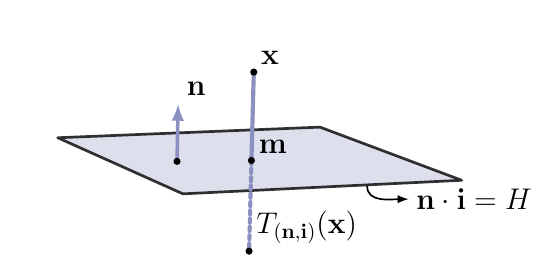}
\vspace{-15pt}
\caption{Reflection across a hyperplane}
\label{fig:reflec}
\end{wrapfigure}

Reflective symmetry can be defined using a hyper-plane parameterized by the surface normal $\nn\in\Re^d, \norm{\nn} = 1$ and a point on the plane $\ii \in\Re^d$: $H(\nn, \ii) = \{\xx \mid (\xx-\ii)^\top\nn = 0\}$.
Given this definition, reflective symmetry is a transformation:
\begin{align}
    T_{(\nn,\ii)}(\xx) = \xx + 2(\proj_{(\nn,\ii)}(\xx) - \xx),
\end{align}
where $\proj_{(\nn,\ii)}(\xx) = \xx + ((\ii-\xx)^\top\nn)\nn$ is the function projecting point $\xx$ to the hyperplane parameterized by the normal vector $\nn$ and a point on the plane $\ii$. 
\noindent The general idea of symmetry detection is to find a hyperplane $(\nn, \ii)$ that defines a reflection applicable to a significant part of the shape.
One way to achieve this is to formulate symmetry detection as finding all hyperplanes under which the result of following optimization problem returns a significant region:
\begin{equation}
    \begin{aligned}
    \label{eq:opt-prob}
        &\text{Find all }(\nn, \ii) \text{ s.t. } 
        M(R^*(\nn, \ii)) > \tau, \\
        &\text{where } R^*(\nn, \ii) = \argmax_{R\subseteq S, \dD(T_{(\nn, \ii)}(R), R) < \eps} M(R) 
    \end{aligned}
\end{equation}

\citet{mitra2006partial} proposes to solve this via a voting algorithm. We provide a brief outline of the 2D algorithm below.
For a given 2D shape, any pair of points $\pp$ and $\qq$ on the boundary of the shape $S$ define a reflective symmetry, represented by the line through the midpoint $\mm=\frac{1}{2}(\pp+\qq)$ with a normal in the direction of $\nn = (\pp-\qq)\norm{\pp-\qq}^{-1}$. 
Now, by sufficiently sampling such point pairs on the shape, we can collect ``votes'' for potential reflective symmetries, $T_i$, in the transformation space. \citet{mitra2006partial} observe that these votes create a density function in the transformation space, and the modes of the such density function correspond to significant symmetries since they are a transformation that is proximal to most votes. In other words, the optimization problem in \cref{eq:opt-prob} can be formulated as a mode seeking procedure of the density defined in the transformation space.

Building on this concept, \citet{mitra2006partial} suggest using mean-shift algorithm \cite{meanshift} to identify clusters associated with significant modes.
Specifically, given a set of candidate transformations $\{T_i\}_{i=1}^n$, the mean-shift algorithm can be thought as first constructing a density field using kernel density estimation:
\begin{align}
    p(T) = \frac{1}{|N(T)|h^d}\sum_{T_i\in N(T)}^n K\paren{\frac{T - T_i}{h}},
\end{align}
where $K$ is an appropriate kernel function, $h$ is the bandwidth, and $N(T)$ contains all transformations near the query $T$.
The algorithm first initializes from any of the candidate transformation $T^{(0)} = T_i$ and performs the following iterations:
\begin{align}\label{eq:meanshift}
    T^{(k+1)} \leftarrow 
    \frac
    {\sum_{T'\in N_k} K((T' - T^{(k)})h^{-1}) T'}
    {\sum_{T'\in N_k} K((T' - T^{(k)})h^{-1})},
\end{align}
where $N_k = N\paren{T^{(k)}}$ is the center of the kernel at the $k^{\text{th}}$ iteration.
This process repeats until the centroid is converged, \ie $\|T^{(k+1)}-T^{(k)}\| < \gamma$ for certain $\gamma$. Once the proposed transformation $T^*$ is found, \citet{mitra2006partial} computes $R^*$ in ~\cref{eq:opt-prob} using a region growing algorithm and filters out $R^*$ if $\mM(R^*)$ is not significant enough.
While the mean-shift voting scheme has shown promising results in detecting varying types and scales of symmetries, they are limited in that it fails under noisy shapes (\ie noisy transformation space) and does not fully leverage modern GPU compute or parallel processing power. 
To tackle such shortcomings, we establish a connection between \citet{mitra2006partial} and the literature state-of-the-art generative models. This connection allows us to leverage advancements in modern computing and deep learning techniques, applying them to the task of symmetry detection.



\section{Method}
\label{sec:method}

In this section, we first establish the connection between Langevin dynamics, as used in diffusion models, and the mean-shift algorithm used in symmetry detection (\cref{sec:flow_match}). Building on the insight derived from this connection, we propose a symmetry detection method using Langevin dynamics to achieve robustness to noise. We also outline the design of the transformation representation, define the distance between reflective symmetries (\cref{sec:transf-space}), and illustrate how to perform Langevin dynamics in this transformation space (\cref{sec:lang_transf}). Finally, we provide a simple algorithm to post-process the resulting modes after Langevin dynamics to extract the final symmetries (\cref{sec:sym-extraction}).

\subsection{Connecting Mean-shift and Langevin Dynamics}
\label{sec:flow_match}
Diffusion models ~\cite{ho2020denoising}  and flow matching ~\cite{lipman2022flow} aim to learn the score function, or the gradient of the log density function.
These score functions can be used with Langevin dynamics (or any other SDE solver) to transport samples from the prior distribution to the data distribution.
Specifically, given empirical data samples $\xX = \{x_i\sim P_{\text{data}}\}_{i=1}^n$,
one way to define the objective of the diffusion model is to learn the score for the following density, which convolves the actual data density with a Gaussian to create a blurry version:
\begin{align}
    P_\sigma(x) = \int P_{\text{data}}(y)\nN(x;y,\sigma^2I) dy 
    \approx \frac{1}{|\xX|}\sum_{i=1}^{|\xX|} \nN(x;x_i, \sigma^2I),
\end{align}
where the RHS is an approximation using empirical samples $\xX$.
This allows us to write the score function~\cite{pabbaraju2023provable, song2021scorebased, cai2020learning, karras2022edm} as the following:
\begin{align}
    \nabla_x \log P_\sigma(x)
    \approx \paren{
        \frac{\sum_{y\in \xX} \nN(x;y,\sigma^2I) \cdot y}{\sum_{y\in \xX} \nN(x;y,\sigma^2I)} 
    - x } \sigma^{-2}.
\end{align}
Once the model learns to approximate this score function, one way to obtain samples is through Langevin dynamics.
Specifically, we can initialize Langevin dynamics with samples from a prior distribution $x^{(0)} \sim P_{\text{prior}}$, and then perform the following iterations:
\begin{align}
    x^{(t+1)} \leftarrow x^{(t)} + \alpha_t \nabla_x\log P_{\sigma_t}(x^{(t)}) + \sqrt{2\alpha_t}\beta_t\epsilon_t, 
\end{align}
where $\epsilon_t\sim\nN(0,I)$ is the noise and $\alpha_t$ is the step size. A key connection here is that if we modify the above-mentioned formula slightly, we are able to obtain the updates of mean-shift in~\cref{eq:meanshift}.
Specifically, if we set $\beta_t = 0$, $\alpha_t =\sigma_t^2$,  and $\sigma_t = h$, then the Langevin updating rule becomes:
\begin{equation}
\begin{aligned}\label{eq:langevin-dynamic}
    x^{(t+1)} 
    &\leftarrow x^{(t)} +h^2\paren{
        \frac{\sum_{y\in \xX} \nN(x^{(t)};y,h^2I) \cdot y}{\sum_{y\in \xX} \nN(x^{(t)};y,h^2I)} 
    - x }/h^2 \\
    &= 
    \frac{\sum_{y\in \xX} \nN\paren{x^{(t)};y,h^2I} \cdot y}{\sum_{y\in \xX} \nN\paren{x^{(t)};y,h^2I}}. 
\end{aligned}
\end{equation}
The update rule above can be thought as mean-shift algorithm using Gaussian kernel $(\ie K(x, y) = 
\exp\left(-\frac{\|x - y\|^2}{h^2}\right))$, but with no stochasticity, infinite neighborhood (\ie $N(x)=\xX$), as well as constant noise value $\sigma_t = h$.
Intuitively, this connection roots in the fact that both Langevin dynamics and mean-shift are trying to perform iterative hill climbing in a density function.

Despite the similarities, there are also differences in the choice of hyperparameters.
For example, adding stochastic noise in Langevin dynamics is generally considered to bring an advantage in sampling quality, allowing us to obtain symmetry modes of varying strengths. A design choice of mean-shift algorithm is to update with the gradient estimated from a finite neighborhood, which makes the update computationally efficient. 
The same hyperparameter choice in Langevin dynamics has similar effects to using a small $\sigma_t$ or truncating the gradients. However, this can be problematic when data is sparse or noisy, potentially creating spurious modes that require post-processing.
These observations lead us to ask: \textit{can diffusion models be applied to perform mode-seeking for symmetry detection?}
Nevertheless, directly applying Langevin dynamics to the space of reflective transformation is challenging because the transformation space has geometries different from Euclidean space, as it contains an $SO(n)$ rotation.
This requires us to construct a space where we can define an appropriate density function, estimate scores, and perform Langevin dynamics.

\subsection{Transformation Space of Reflective Symmetry\label{sec:transf-space}}

As we directly walk on the transformation manifold, choosing the optimal transformation representation was of utmost importance. 
Previous works have explored representing transformations as Lie groups \cite{shi2016symmetry} and Euclidean transformation groups \cite{mitra2006partial}, yet these representations are often much higher dimensional -- typically in $\Re^7$. 
Such complexity makes identification of high-density regions in the space challenging. 

A good space for diffusion model should be bounded and easy to compute. Therefore, we resort to the Hesse normal form utilized in Hough transforms \cite{Duda1972UseOT} to represent the hyperplane that defines a reflective symmetry. Hough transform \sloppy 
\begin{wrapfigure}{r}{0.275\textwidth}

\centering
\includegraphics[width=\linewidth]{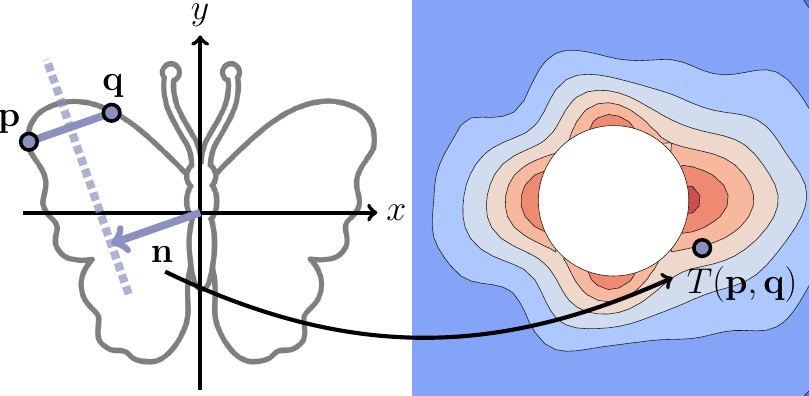}
\vspace{-18pt}
    \caption{Transformation space construction}
    \label{fig:sym_extraction}
\end{wrapfigure}

 \noindent represents each hyperplane $\{\xx|(\xx-\ii)^T\nn=0\}$ using a rotation parameterized by the surface normal vector $\nn\in\Re^d$, oriented away from the origin, and a scalar $l\in\Re$ denoting the shortest distance to the plane from the origin. Given this representation $(\nn, l)$, the reflective plane can be written as:
$
    \{\xx| \xx^T\nn = l\}.
$
A convenient property of Hough transform is that both $\nn$ and $l$ are bounded as long as the input shape $S$ is bounded.
Moreover, it's easy to compute the potential reflective plane for a pair of points in the shape $(\pp, \qq) \in S$:
\begin{align}
    \nn(\pp, \qq) = (\pp-\qq)\norm{\pp-\qq}^{-1},\quad l(\pp,\qq) = \frac{1}{2}(\pp+\qq)^\top n(\pp,\qq)
\end{align}
Such computation is very efficient and parallelizable, allowing us to scale to a large number of points.
Instead of doing walks in the concatenate space $(\nn,l)$, which has one additional degree of freedom than the actual space of reflective symmetry, we embed the symmetry into $\Re^d$ using the following:
\begin{align}
\label{eq:sym_emb}
    T(\pp,\qq) = n(\pp,\qq) \cdot \paren{\operatorname{sign}(l(\pp,\qq)) \cdot k + l(\pp,\qq)}
\end{align}
where $k$ is the hyperparameter denoting the radius of the invalid region. This parameter is introduced to deal with ambiguities when $l(\pp,\qq)=0$, where multiple symmetry planes are mapped to the origin. Specifically, we shift all planes in the direction of the normal by $k$. This representation is numerically stable in most cases except for when $\pp=\qq$, which we discard as they do not create a valid vote for reflective symmetry. Each reflective hyperplane thus gets mapped to a single point in the transformation space (\cref{fig:sym_extraction}). \begin{figure}[t]
    \includegraphics[width=\linewidth]{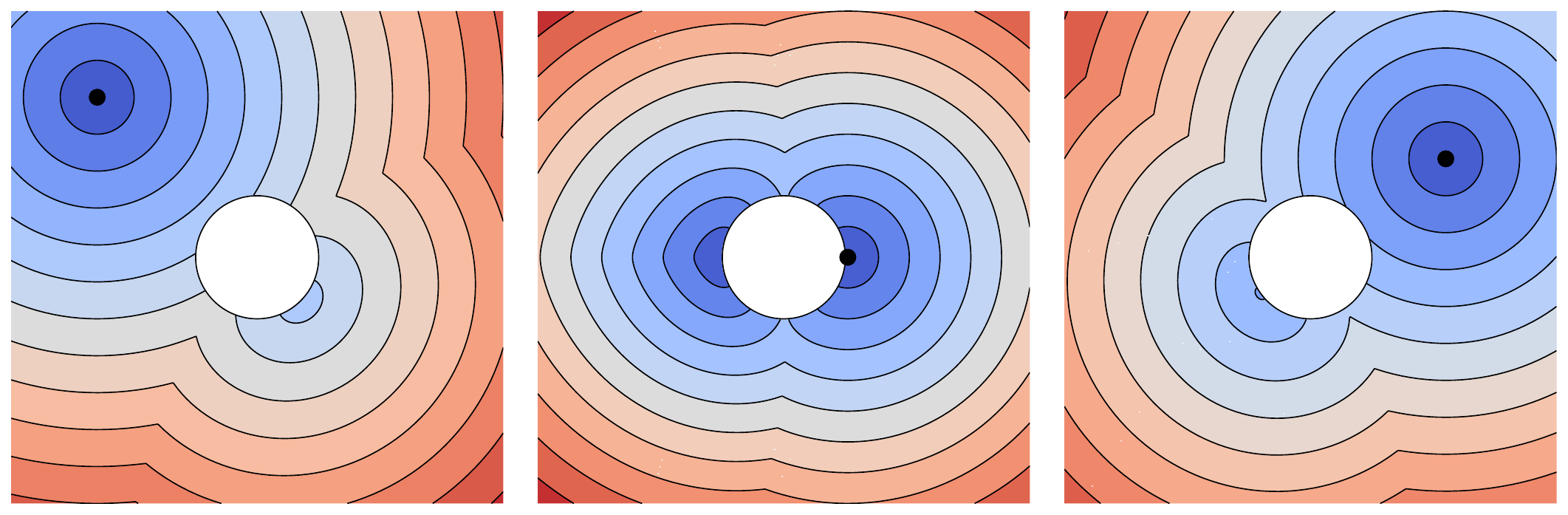}
    \vspace{-2em}
    \caption{
       Geodesic distance to the black point in transformation space. Blue indicates proximity, and red indicates larger distance. Shortest distance can be computed traveling through the origin as well. 
    }
    \label{fig:geodesic}
    \vspace{-15pt}
\end{figure}

\subsection{Langevin Dynamics in Transformation Space}
\label{sec:lang_transf}
One challenge of performing the walk in the transformation space defined as $T(\pp,\qq)$ is that Euclidean distance in the space does not reflect the actual distance between symmetry planes.
This is due to the fact that reflective symmetry contains $SO(n)$ rotation symmetry.
Fortunately, the score function can be computed in a Riemannian manifold where only distances are defined~\cite{chen2024flow,huang2022riemannian}.
Below, we first cover the background of score estimation in a Riemannian manifold, which enables us to compute the geodesic function for the reflective transformation space. Next, using the score function derived from the geodesic, we define the process of walking in the transformation space.
\paragraph{Langevin Dynamics in Riemannian Geometry.}
Let $\mM$ be a Riemannian manifold where we define a premetrics measuring the closeness of points $\dd:\mM\times\mM\to\Re$ such that $\dd$ is (1) non-negative (\ie $\dd(x,y)\geq 0, \forall x,y\in \mM$), (2) positive (\ie $\dd(x,y) = 0 \text{ iff } x=y$), and (3) non-degenerate (\ie $\nabla_x\dd(x,y) \neq 0 \text{ iff } x\neq y$).
Given the premetrics function, \citet{chen2024flow} suggest that one way to compute the gradient of the probability density flow is using the following formula:
\begin{align}
    g_\sigma(x) &= \int_{\mM}\dd(x, y)\frac{\nabla_x \dd(x, y)}{\norm{\nabla_x\dd(x,y)}^2}\frac{p_\sigma(x|y)}{p_\sigma(x)}P_{\text{data}}(y)~d\text{vol}_y,  
\end{align}
where $p_\sigma(x|y)$ is a Gaussian distribution on the geodesic distance:
$
    p_\sigma(x|y) \propto \exp\paren{-\dd(x,y)^2\sigma^{-2}}.
$
The gradient field of the probability density flow can be used in place of the score function in Langevin dynamics. This means as long as we are able to create a premetrics (or geodesic distance) for the transformation space $\mM$ to reflect the closeness of reflective symmetries, we are able to compute $g_\sigma$ to perform Langevian dynamics.

\paragraph{Distance of Reflective Symmetry.}
Intuitively, the distance between two reflective symmetries $T_1, T_2$ should capture how hard it is to transform one hyperplane to the other. 
A seemingly straightforward approach is to directly compute the Euclidean distance between $T_1$ and $T_2$.
While this gives some measure of geodesic, it is problematic near the origin because when $l=0$, both $\nn$ and $-\nn$ represent the same hyperplane, yet $(0, \nn)$ and $(0, -\nn)$ is far away in our representation (\cref{eq:sym_emb}).
To address this issue, we propose the following geodesic distance:
\begin{align}
    \dd(x,y) = \min\begin{cases}
        \min_z \norm{x-z} + \norm{y+z} \\
        \min_r \int_{0}^1 \operatorname{valid}(r(t))|r'(t)|dt 
    \end{cases},
\end{align}
where $\norm{z}=k$, $r(0) = x$, $r(1)=y$, and $\operatorname{valid}(x) = 1 + \infty\cdot\mathbf{1}(\norm{x} > k)$ marks the valid region of the path allowed.
Intuitively, the first term captures the shortest path if the hyperplane moves past the origin. 
The second term captures the shortest path if we do not go through the origin. We visualize this geodesic distance at different points in ~\cref{fig:geodesic}. We implement the geodesic with JAX and compute the gradient $\nabla_x \dd(x, y)$ using automatic differentiation.


\paragraph{Stepping in the Transformation space.}
Given the definition of the vector field $g_\sigma$ which maps a point in the geodesic to the tangent space, we can replace the score function in~\cref{eq:langevin-dynamic}. However, naively doing so will lead to an issue where the points might end up in invalid areas that don't represent any valid reflective symmetry plane (\eg $\|x^{(t)}\|<k$). 
This occurs if the gradient field $g_\sigma$ is trying to move points across the origin to find symmetries with more votes.
One way to counter this is to properly define the behavior of moving a hyperplane across the origin with certain velocity.
Specifically, suppose $T$ is a point with $l=0$ and normal $\nn$ (so it's on the sphere with radius $k$).
If we would like to move $T$ with velocity $v_d$ and $v_n$, this is equivalent to moving $-T$ (\ie $l=0$ and $-\nn$) with $v_d$ and $v_n$.
If we encounter such a case during the walk, we consider moving both $T$ and $-T$, then only send the point to the valid region.

\begin{wrapfigure}{r}{0.18\textwidth}
\includegraphics[width=\linewidth]{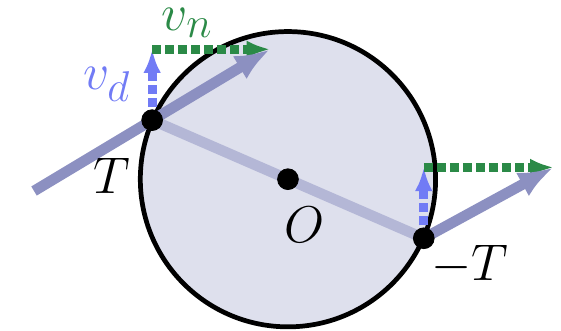}
\vspace{-18pt}
\caption{Stepping in the transformation space. We move both $T$ and $-T$, and only send the point to the valid destination.}
\label{fig:step}
\end{wrapfigure}

\noindent This operation is performed both during the gradient step (\ie walking along $g_\sigma$) as well as the noise perturbation step (\ie walking along $\epsilon$). We illustrate the full Langevin dynamics trajectory as well as the corresponding hyperplane in the shape space upon convergence in ~\cref{fig:langevin}. We show the detailed implementation and hyperparameter setup in the appendix.
 

\subsection{Symmetry Extraction\label{sec:sym-extraction}}
After sufficient steps of Langevin dynamics, randomly initialized points at the beginning of the walk will eventually converge at the local maxima of the density field if guided properly. However, this still gives us a set of extremely narrow distributions centered at each mode. In order to extract a single symmetry for every peak, we perform Density-Based Spatial Clustering of Applications with Noise (DBSCAN), a density-based clustering algorithm that groups points based on neighborhood density \cite{dbscan,schubert2017dbscan,khan2014dbscan}. After Langevin we already expect the modes to be well-separated and dense, thus, DBSCAN is only utilized to identify the centroids and to remove potentially remaining low-density points. Therefore, we enforce DBSCAN with a very high density threshold and a small distance threshold. Once the centroid at each local maxima is identified, we map it back to shape space to find the corresponding symmetry. We provide experiment results quantifying the impact of DBSCAN in the appendix.
\begin{figure}[t]
\centering
\includegraphics[width=\linewidth, height=125pt]{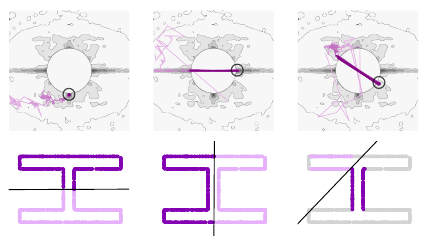}
\vspace{-2.5em}
\caption{
Visualizing Langevin dynamics trajectory. For each converged mode, we show the full trajectory (top row), as well as the corresponding symmetry in the shape space (bottom row). Note that due to our Langevin construction, we directly jump across the invalid region as opposed to stepping within it. When converged modes are near the boundary of the invalid region ($l \approx k$), we see oscillations through the origin as the two modes are equivalent ($H(0, \nn) = H(0, -\nn)$).}
\vspace{-1em}
\label{fig:langevin}

\end{figure}

\subsection{Datasets}
We evaluate our symmetry detection pipeline on both 2D and 3D shapes. In order to evaluate the performance of the proposed symmetry detection pipeline, there has to be a good amount of symmetries present in the dataset; yet common baselines such as MNIST datasets~\cite{lecun1998gradient} do not contain much symmetries. Therefore, for 2D shapes, we construct a dataset of English alphabet letters of varying fronts from Google Fonts~\footnote{\url{https://fonts.google.com/}}. We extract 10 fonts for each letter, and extract the contour. For 3D shapes, we utilize 20 selected object meshes from ShapeNetCore~\cite{chang2015shapenet} and Adobe Assets. We normalize all shapes by shifting the center of mass to the origin and scaling them such that all points are within $[-1,1]^d$. To assess the noise-robustness of the method, we introduce varying degrees of noise to the constructed dataset. For 2D, we directly add scaled Gaussian noise $\alpha\cdot \nN(0,1)\text{ for }\alpha = [0 \%$, $1 \%$, $3 \%$, $5 \%]$ to all sampled points on the normalized shape. For 3D, we add up to 3\% noise to all vertices in the direction of the normal. This results in multiple variations of each shape at different noise levels.

However, a key challenge remains in data annotation, as there exists no prior symmetry ground truth. Therefore, we develop a pipeline for generating ground truth symmetry annotations on collected shapes. We first take a brute force approach and sample point pairs to construct a symmetry transformation space. For each proposal from a point pair we calculate the number of points that would vote for this proposal, and only keep the proposals that have more than a threshold number of votes. We then apply clustering to group similar transformations together to collect a set of symmetry proposals. Thresholds in this process are set fairly low such that we always propose more symmetries than there actually are. Given these loose proposals, a human annotator can either decide whether to include the proposed symmetry in the ground truth set or not.

\subsection{Evaluation metrics}

In our analysis, we employ precision and recall as our primary evaluation metrics. By comparing the set of predicted symmetries and the set of ground truth symmetries, the two metrics could be directly obtained. During our evaluation, we consider two symmetries to be the same if they are within some $\delta$-radius of each other in the transformation space. However, due to the labor-intensiveness of the human-labeling of ground truth data, we propose two additional metrics to evaluate the quality of a predicted set of symmetries: association and compression rate. 

To obtain the association measure, we first define the association of a point with a particular symmetry as the condition where the projection of the point through the symmetry falls within an epsilon radius of another point. Then for each symmetry, we calculate the proportion of points satisfying this criterion. We then determine the distribution of these proportions across the set of symmetries. The association measure of the symmetry set is quantified as the integral of this distribution, representing the area under the curve across all ratios of total number of points. This metric serves as a precision measure, capturing the degree to which the set of symmetries accurately associate points of the shape within a defined tolerance.

On the other hand, for the compression measure, we use a heuristic-based compression algorithm to apply symmetries for reducing the number of points required to represent a shape, given the suspected NP-hard nature of an optimal compression algorithm. This algorithm iteratively applies the best symmetries to minimize the representation space, and the resultant compression ratio is calculated by dividing the original memory needed to store all points by themselves with the memory required for the compressed object. Note that the compressed object includes spatial locations of retained points, the symmetries utilized, and the sequence of their application. A lower compression ratio indicates a higher efficacy and relevance of the symmetries used, which serves as a proxy for recall.

\label{sec:shape_refine}

\section{Results and Applications}
\label{sec:results}

\begin{wraptable}{r}{0.26\textwidth} 

\centering
\caption{F1 score for 3D baselines}
\vspace*{-0.8em}
\setlength{\tabcolsep}{2pt} 
\renewcommand{\arraystretch}{1.0} 
\begin{tabular}{lccc}
\toprule
\text{Method$\backslash$Noise} &\textbf{0\%} & \textbf{1\%} & \textbf{3\%}  \\
\midrule
\text{Mitra} & 0.590 & 0.495 & 0.347 \\
\text{PRS-Net} & 0.249 & 0.126 & 0.227 \\
\text{E3Sym} & 0.322 & 0.193 & 0.249 \\
\textbf{Ours} & \textbf{0.745}& \textbf{0.677} & \textbf{0.440} \\
\bottomrule
\end{tabular}
\label{tab:quant_3d}
\end{wraptable}

In this section, we provide results indicating the robustness of our method, which can identify reflective symmetries of different strengths even on noisy shapes. We first show demonstrative examples of detected symmetry planes, evaluate our method quantitatively with other baselines, and provide potential downstream applications.
\begin{figure}[t]
\centering
    \includegraphics[width=\linewidth, trim={0 100 0 0}, clip]{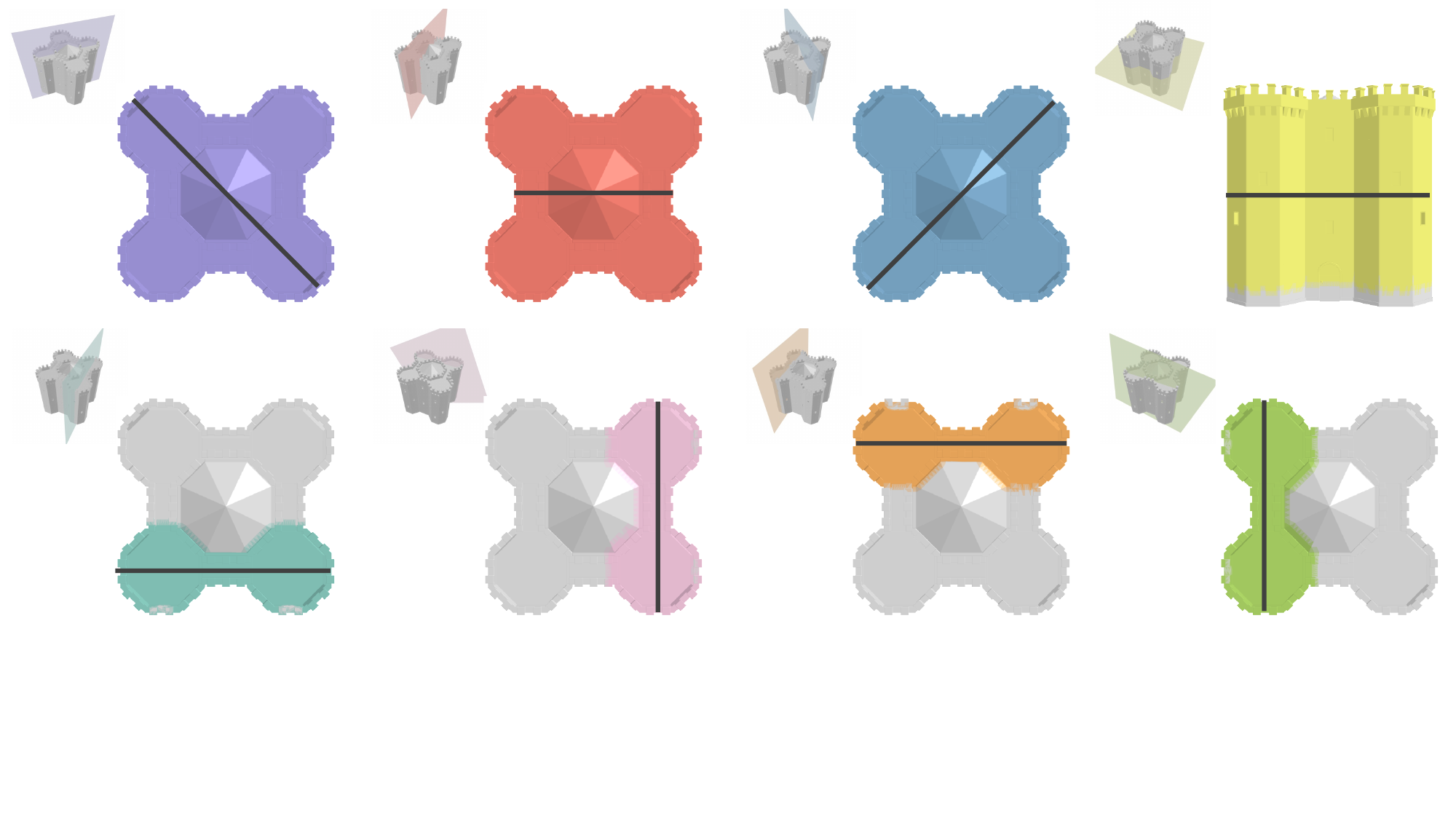}
    \vspace{-3em}
    \caption{Symmetry groups found within a castle wall. The shape patches that support the detected symmetries are colored the same as the corresponding plane. We identify both local and global reflective symmetries.}
    \label{fig:wallsym}
    \vspace{-1em}
\end{figure}

\begin{figure}[t]
\vspace{-5pt}
\centering
\includegraphics[width=\linewidth]{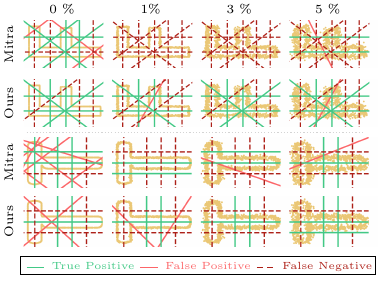}
\vspace{-20pt}
\caption{
Qualitative comparison with \citet{mitra2006partial} for 2D. We visualize the input 2D font shapes and the predicted reflective symmetries for varying levels of noise. We show that our solution is more robust to noise, capable of detecting key symmetries even at high levels of noise.
}
\label{fig:quali_2d}
\end{figure}
\begin{figure}[t]
\vspace{-10pt}
\centering
\includegraphics[width=\linewidth]{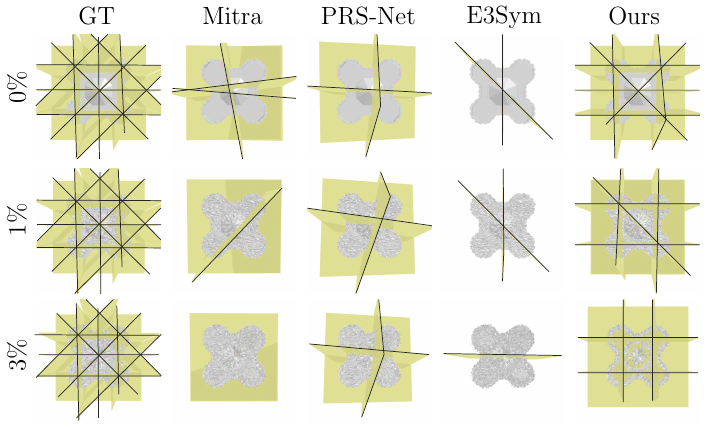}
\vspace{-20pt}
\caption{Qualitative comparison with baselines for 3D. All detected symmetries are shown for 0, 1, 3\% noise. Our method is capable of detecting both local and global symmetries accurately even at high noise levels.}
\label{fig:quali_comp}
\vspace{-15pt}
\end{figure}

\subsection{Evaluation on Symmetry Detection Performance}
\begin{table*}[ht]
    \centering
    \caption{Quantitative results for 2D symmetry detection.}
    \vspace{-1em}
    \setlength{\tabcolsep}{2pt} 
    \renewcommand{\arraystretch}{1.0} 
    \begin{tabular}{@{}lcccccccccccc@{}}
        \toprule
         & \multicolumn{4}{c}{F1 $\uparrow$} & \multicolumn{4}{c}{Compression $\downarrow$ } & \multicolumn{4}{c}{Association $\uparrow$} \\
        \cmidrule(lr){2-5} \cmidrule(lr){6-9} \cmidrule(lr){10-13}
        Methods & 0\% & 1\% & 3\% & 5\%& 0\% & 1\% & 3\% & 5\%& 0\% & 1\% & 3\% & 5\%\\
        \midrule
        \hspace{3mm}Mitra & \textbf{0.602} &0.208 & 0.287 & 0.347 & 0.483 & 0.793  & 0.783  & 0.757 & 0.366  & 0.303  & 0.234  & 0.246\\
        \hspace{3mm}\textbf{Ours} & 0.576 &\textbf{0.540}& \textbf{0.481}& \textbf{0.443}& \textbf{0.389}& \textbf{0.413} & \textbf{0.478}& \textbf{0.530}& \textbf{0.476} &\textbf{0.484} & \textbf{0.477}    & \textbf{0.459} \\

        \bottomrule
    \end{tabular}
\label{tab:quantitative_2d}
\end{table*}

\begin{figure*}[ht]
\centering
\vspace{1em}
\includegraphics[width=\linewidth, height=320px]{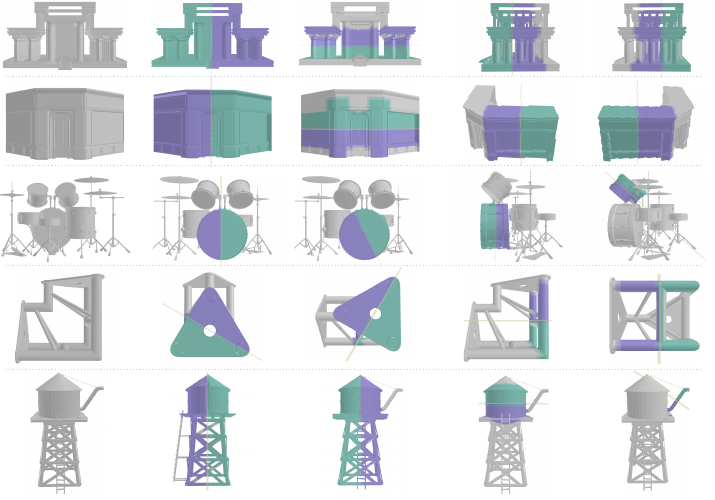}
\vspace{-2em}
\caption{Qualitative results for 3D.
We visualize the input 3D shape and selected reflective planes from our detection results. We only highlight the largest connected region that agree to the symmetry, and color each side of the reflective plane differently. Results indicate that our method is capable of detecting symmetries of varying strengths on a wide array of shapes.
}
\label{fig:quali_3d}
\end{figure*}
\begin{figure}[t]
\centering
    \includegraphics[width=\linewidth]{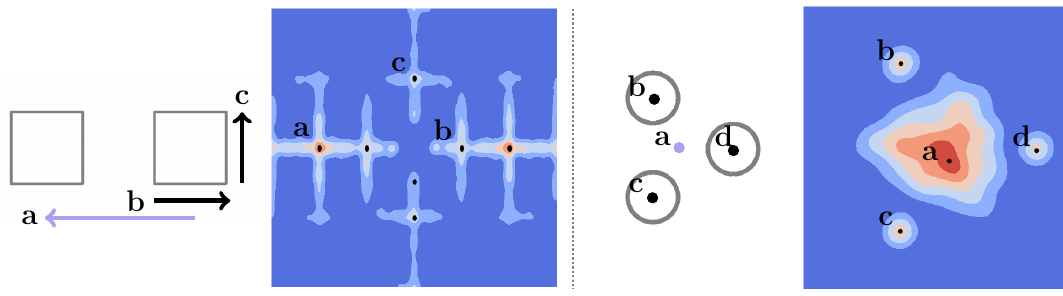}
    \vspace{-2em}
    \caption{We show that our pipeline can be extended to translation and rotational symmetries. For translation, we identify global negative (a) and local positive shifts (b, c). For rotation, we identify global (a) and local rotational axis (b, c, d). The transformation space (left) and the identified modes (right) for both symmetry types are shown. 
    }
    \vspace{-1em}
    \label{fig:addsym}

\end{figure}

\begin{figure*}[ht]
\centering
    \hspace{1cm} Input \hspace{5.6cm} Mitra \hspace{5.6cm} Ours\\
    \includegraphics[width=\linewidth, trim={0.4cm 1.2cm 0.3cm 0.3cm}, clip]{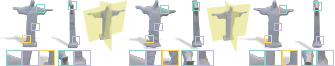}
    \vspace{-2em}
    \caption{Applications in symmetry fixing for noisy shapes. We take an imperfect mesh generated from a single view image, then symmetrize the shape based on the detected symmetry planes. Shapes fixed with our detection is of higher quality than with \citet{mitra2006partial} as our method is more robust to noise. 
     }
    \label{fig:fixing}
    \vspace{-1em}
\end{figure*}

\label{sec:results_2d}

We show quantitative comparisons with neural baselines and \citet{mitra2006partial} for 3D (\cref{tab:quant_3d}) and 2D shapes (\cref{tab:quantitative_2d}). Results indicate that our method outperforms the baseline methods for F1 score, compression, and association, especially by a large margin at high noise levels. 
This is potentially because our method works under a Gaussian-smoothed transformation space, where Langevin dynamics is utilized to help travel through the potentially noisy parameter space and extract the modes, bypassing certain noisy information. On the other hand, \citet{mitra2006partial} directly operates under the noisy transformation space and further uses stochastic clustering to extract the modes. As both procedures are not robust to noise, this likely contributes to its weaker performances on noisy shapes. It is also worth noting that our method has significantly improved compression and association rates across all noise levels, which further demonstrates its advantages in potential downstream applications.


We show qualitative results of our symmetry detection algorithm in comparison with baselines for 2D and 3D (\cref{fig:quali_2d}, \cref{fig:quali_comp}), and results on additional 3D shapes (\cref{fig:quali_3d}). While we miss smaller symmetries as noise is introduced, our method retains higher recall across all levels when compared to baseline methods. Particularly, we see that we are able to constantly identify the biggest modes (i.e. global reflective symmetry) throughout. Whereas previous neural methods were limited to global symmetries, we identify symmetries of varying strengths on a wide range of unseen shapes.

\begin{figure}[t]
\centering
    \includegraphics[width=\linewidth, height=150px ]{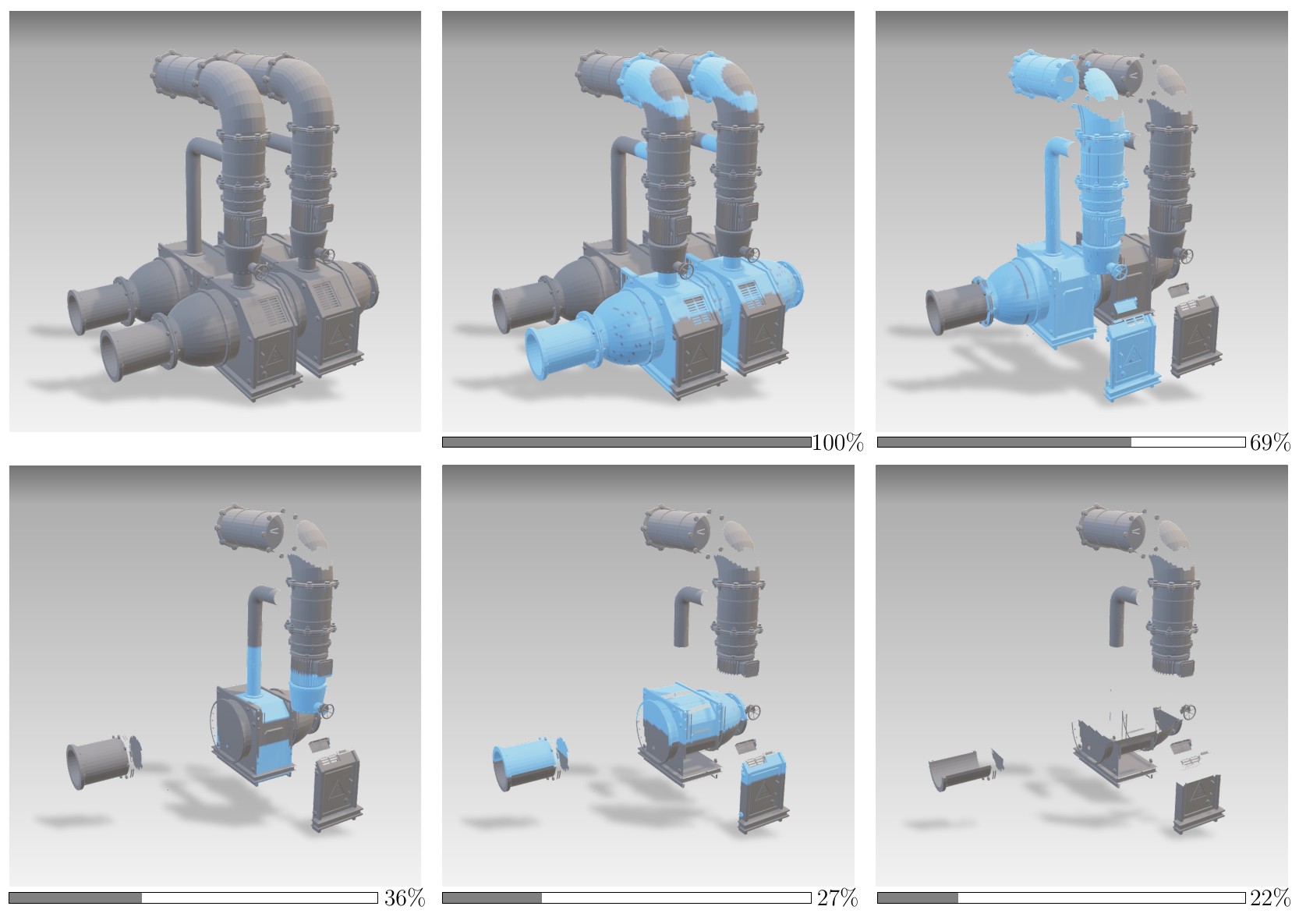}
    \vspace{-2em}
    \caption{Applications in shape compression. We take a shape with symmetries at many different scales, and perform sequential compression based on the detected modes. Region to be compressed is colored blue at each stage. 
    }
    \label{fig:compress}
\vspace{-1em}
\end{figure}

\subsection{Applications}
\label{sec:applications}
In this section, we demonstrate potential downstream applications utilizing the detected symmetries. We first show an extension to different types of symmetries - translation and rotation, then demonstrate applications in shape fixing and compression. 

\paragraph{Extension to additional symmetry types}
As our approach is training-free, extending the pipeline to other symmetry types is possible as long as a representative and bounded transformation space can be constructed. We show proof-of-concept examples of extensions to rotation and translation symmetries, where we only modify the transformation space we walk on (\cref{fig:addsym}). For constructing the transformation space for translation, we directly extract vector displacement for every point pair (instead of the Hough transform representation). For rotational symmetries, we utilize a special case of the Cartan-Dieudonn\'e theorem, which states that every rotation in SO(n) is a composition of at most n-reflections about hyperplanes. Given two reflective planes, the rotational axis corresponds to the line of intersection between two planes, where the angle of rotation is twice the angle of intersection. With this idea, we directly extend the reflective transformation space to rotational space. 
\paragraph{Extension to shape fixing}
Utilizing the method's robustness to noise, we show potential applications to downstream shape fixing tasks (\cref{fig:fixing}). We implement the shape symmetrization pipeline proposed in \citet{mitra2007symmetrization}. We demonstrate that given an imperfect shape generated from SOTA image-to-mesh model \cite{xu2024instantmeshefficient3dmesh}, we can accurately detect symmetry planes and fix the shapes accordingly. As our method can be directly applied to unseen noisy shapes, it could be further utilized in correcting the output of 3D generative models. 
\paragraph{Extension to shape compression}
Another downstream application for symmetry detection is in shape compression. We perform sequential model compression based on the detected local and global reflective planes (\cref{fig:compress}) with the pipeline. Using the first four significant modes, a reduction to only 22\% of the original model size is achieved by removing redundant shape patches. This demonstrates the potential of symmetries as a complexity-reducing concept to simplify shapes.


\section{Future Work and Limitations}
\label{sec:futurework}
In this work, we bring a novel perspective to the symmetry detection problem and propose a noise-robust algorithm. Building on classical approaches, we introduce modern concepts from generative models, particularly Langevin dynamics, to address prior shortcomings. However, there are a few limitations that are worth exploring in the future.
\paragraph{Neural gradient field}
We show that with sufficiently sampling the shape space, we are able to construct a transformation space that is representative of the symmetries present within the shape. Yet, the quality of the transformation space as well as the gradient field constructed from this space is heavily dependent on the sample size. Therefore, one may learn this gradient field to better capture the transformation space without sampling.
\paragraph{Collecting ground truth evaluation data}
As there exists no prior symmetry dataset, the evaluation of the method is greatly limited by the lack of fine-grained symmetry annotations. In this work we propose a semi-supervised method to generate a base set of ground truth data, although expanding this to a larger dataset that captures different types of symmetries will help perform a more comprehensive evaluation.
\paragraph{Covering a larger symmetry space}
While we show that our pipeline is capable of identifying reflective symmetries both in partial and global scales, there are other types of symmetries that we did not cover extensively in the scope of this work. Further extending the work to continuous symmetries (e.g. rotational and helical surfaces), translational symmetries, or a combination of multiple symmetry types would be a promising next step.

\section{Conclusions}
\label{sec:conclusion}
We have introduced a novel symmetry detection algorithm based on Riemannian Langevin dynamics. We formulate the transformation space as a score field, on which we perform mode-seeking with Langevin dynamics.

Our experiments validate the robustness and generalizability of the approach, showing improved mode detection even in the presence of noise. The proposed method is capable of detecting both partial and global reflective symmetries in 2D and 3D shapes, which addresses limitations with previous neural approaches. Finally, we show that our detection method generalizes easily to a wide range of shapes and can be integrated into various downstream tasks, highlighting the practical value of our algorithm.

\begin{acks}
This work was supported by ARL grant W911NF-21-2-0104, Vannevar Bush Faculty Fellowship, and Azrieli Foundation Early Career Faculty Fellowship. We would also like to thank Eric Chan, Marcel R\o d, Lucia Zheng, Chenglei Si, Zoe Wefers, George Nakayama, and Maolin Gao for all the helpful discussions and feedback. 
\end{acks}

\bibliographystyle{ACM-Reference-Format}
\balance
\bibliography{ref}

\clearpage
\appendix

\onecolumn

\setcounter{table}{0}
\renewcommand{\thetable}{A\arabic{table}}
\renewcommand{\thefigure}{A\arabic{figure}}
\setcounter{figure}{0}
\section*{\textbf{APPENDIX}}

\section{\textbf{Method Details}}
We provide algorithm blocks to better illustrate the Langevin dynamics process on the transformation manifold. See \cref{alg: lang} for the overall Langevin dynamics setup, \cref{alg: geodesic} for geodesic distance compute, and \cref{alg: walk} for walking on the manifold. Please refer to \cref{sec:hyper} for detailed hyperparameter setup. \\

\SetKwComment{Comment}{/* }{ */}
\begin{algorithm}
\DontPrintSemicolon
    \textbf{Require: }$\text{Noise levels } \{\sigma_i\}^k_{i=1} \text{; Step size } \alpha\text{; Number of steps }T$\;
    Initialize $x_0$\;
    \For{$i \gets 1$ to $k$}{
        \For{$t \gets 0$ to $T-1$}{
            $\epsilon_t \sim \mathcal{N}(0,1)$\;
            $x'_{t+1} \gets x_t + \frac{\sqrt{\alpha}\sigma_i\epsilon_t}{\sigma_k}$\;
            $g_\sigma(x) \gets \text{jax.grad(}\FuncSty{geodesic}(x)\text{)}$\;
            $x_{t+1} \gets \FuncSty{walk}(x'_{t+1}, \frac{\alpha\sigma_i^2}{2\sigma_k^2}g_\sigma(x'_{t+1}, \sigma_i))$\;
        }
        $x_0 \gets x_T$\;
    }
\caption{Langevin Dynamics.}
\label{alg: lang}
\end{algorithm}
\begin{algorithm}
    \DontPrintSemicolon
    \textbf{Require: }\text{x; y; Radius of the invalid region $R$}
    \SetKwFunction{FMain}{\text{geodesic}}
    \SetKwProg{Fn}{Function}{:}{}
    
    \Fn{\FMain{$x$, $y$, $R$}}{
        $x_{valid}, y_{valid} \gets \neg \text{inside\_invalid}(x, R), \neg \text{inside\_invalid}(y, R)$\;
        
        \eIf{$x_{valid} \land y_{valid}$}{
            $d_{through} \gets \text{distance\_jump\_invalid}(x, y, R)$\;
            $d_{outside} \gets \text{distance\_outside\_invalid}(x, y, R)$\;
            \Return $\min(d_{through}, d_{outside})$\;
        }{
            \Return $-1$\;
        }
    }
    \caption{Computing the Geodesic Distance.}
    \label{alg: geodesic}
\end{algorithm}
\begin{algorithm}
    \DontPrintSemicolon
    \textbf{Require: }\text{x; Gradient $g$; Radius of the invalid region $R$}
    \SetKwFunction{FMain}{\text{walk}}
    \SetKwProg{Fn}{Function}{:}{}
    
    \Fn{\FMain{$x$, $g$, $R$}}{
        $\text{target} \gets x + g$\;
        $\text{inside} \gets \textbf{True} \text{ if } \| \text{target} \| < R; \text{ else }\textbf{False}$\;
        
        $t_M \gets -x \cdot g$\;
        $M \gets x + t_M * g$\;

        $\theta_{rot} \gets \arccos(\frac{\|M\|}{R})$\;

        $Y_{pos}, Y_{neg} \gets \text{rotation\_matrix}(\pm\theta_{rot}) \cdot \frac{M}{\|M\|} \cdot R$\;

        $T_{Ypos}, T_{Yneg} \gets (Y_{pos}, Y_{neg} - x) \cdot g$\;


        $Y \gets x + \min(T_{Ypos}, T_{Yneg}) * g$\;

        $v_y \gets \text{target} - Y$\;
        $v_N \gets (v_y \cdot Y) \cdot Y / R$\;
        $\theta_{sign} \gets \text{sign}(\frac{v_y - v_N}{\|v_y - v_N\|}\times\frac{Y}{R})$\;
        $Y_{refl} \gets -\text{rotation\_matrix}(\frac{\|v_y - v_N\|}{R} * \theta_{sign}) \cdot Y$\;
        $\text{target}_{refl} \gets Y_{refl} + \frac{Y_{refl}}{\|Y_{refl}\|} \cdot \|v_N\|$\;
        \eIf{\text{inside}}{\Return $\text{target}_{refl}$ }{\Return $\text{target}$}
    }
    \caption{Walking in the Transformation Space.}
    \label{alg: walk}
\end{algorithm}
\vspace{-10em}

\clearpage
\section{\textbf{Experiment Details}}
\label{sec:hyper}

The parameters we tune for transformation space construction are: $k$ (the radius of the invalid region) and the number of sampled points. The parameter $k$ is introduced to deal with ambiguities when \(l(p,q)=0\), where multiple symmetry planes are mapped to the origin. Specifically, we shift all planes in the direction of the normal by $k$ (Eq. 12). We find that 50K point pairs sampled from both 2D and 3D shapes are sufficient to construct a dense transformation space to capture local modes. 

\begin{wraptable}{r}{0.38\textwidth}
\vspace{-1em}
    \centering
    \caption{Runtime analysis}
    \vspace{-1em}
    \begin{tabular}{cccccc}
    \toprule
    \multirow{2}{*}{} & \multicolumn{4}{c}{Number of steps} \\
    \cmidrule{2-5}
    &1K & 5K & 10K & 50K \\
    \midrule
    F1 & 0.759 & 0.813 & 0.774 & \textbf{0.857} \\
    
    Runtime & 29.3\text{s} & 100\text{s} & ~4\text{min} & ~15\text{min} \\
    \bottomrule
    \label{tab:runtime}
    \end{tabular}
    
\end{wraptable}

For mode seeking, we utilize the annealed version of Langevin dynamics \cite{song2019estimating}. We begin with random samples and perform recursive updates toward the gradient while gradually decreasing noise. The gradient is computed from the kernel-approximated geodesic distances \cite{chen2024flow, huang2022riemannian}. This requires tuning kernel size, step size, and number of steps. A smaller kernel size allows finer modes to be discovered but may require more steps for convergence, while a larger kernel size may overlook these finer modes. With hyperparameter tuning, we achieve the best performance with: kernel size = 0.025, $k$ = 0.3, step size = 0.06 for 2D, and kernel size = 0.08, $k$ = 0.5, step size = 0.02 for 3D. We use 200 points for Langevin initialization and take 50K steps for both. We show the effects of altering such parameters below, where we profile end-to-end runtime for varying number of steps, for which we measure the wall-clock time on a 12G RTX 2080 GPU (\cref{tab:runtime}). We also quantify the importance of the hyperparameter $k$ as well as kernel size (\cref{tab:hyp}). We report the F1 score for all.\\

\begin{table}[htbp!]
\centering
\vspace{2pt}
\caption{Hyperparameter setup}
    \centering
    \begin{tabular}{cccccccccc}
    \toprule
    \multirow{2}{*}{} & \multicolumn{4}{c}{$k$} & & \multicolumn{4}{c}{Kernel size} \\
    \cmidrule{2-5} \cmidrule{7-10}
    & 0.3 & 0.5 & 0.8 & 1 & & 0.2 & 0.1 & 0.08 & 0.05 \\
    \midrule
    F1 & 0.733 & \textbf{0.857} & 0.710 & 0.733 & & 0.700 & 0.846 & \textbf{0.857} & 0.560 \\
    \bottomrule
    \label{tab:hyp}
    \end{tabular}
\end{table}

\section{\textbf{Ablation studies}}

\begin{wraptable}{r}{0.40\textwidth} 
\centering
\caption{Ablation study on the impact of DBSCAN}
\begin{tabular}{@{}lcccc@{}} 
\toprule
& \textbf{0\%} & \textbf{1\%} & \textbf{3\%} & \textbf{5\%} \\
\midrule
Mitra & \textbf{0.602} & 0.208 & 0.287 & 0.347\\
Mitra + DBSCAN & 0.517 & 0.182 & 0.120 & 0.097\\
\textbf{Ours} & 0.576 & \textbf{0.540} & \textbf{0.481} & \textbf{0.443} \\
\bottomrule
\end{tabular}
\label{tab:dbscan_ablate}
\end{wraptable}

As we assume the converged modes after Langevin steps to be well-seperated and dense, we only utilize DBSCAN as a post-processing step to extract final mode centroids. With this assumption, we enforce it with a very high density and a small distance threshold. In order to confirm that we are not relying on DBSCAN for clustering, we perform an ablation study quantifying the impact of DBSCAN. We replace mean-shift clustering in \citep{mitra2006partial} with DBSCAN and report the F1 score for all noise levels. We determine the hyperparameters for DBSCAN following the approaches in Sander et al. [1998]. We observe that DBSCAN struggles with noisy shapes even more so than mean-shift, likely due to its reliance on fixed-density thresholds. 

\section{\textbf{Robustness to sparse reconstruction}}
We acknowledge that directly adding noise to the shape might not be realistically capturing noises encountered in real-life applications. Therefore, we evaluate the method further on sparsely-reconstructed point clouds, which are generated from back-projecting 4, 6, and 8-views randomly taken from the input 3D shape. We show that the pipeline is capable of consistently detecting modes even in sparse conditions. \\
\begin{figure*}[ht]
\centering
   \hspace{-2cm}4 views \hspace{1.3cm} 6 views \hspace{1.3cm} 8 views \hspace{3.5cm} Top 3 planes\hspace{7cm} \\
    \includegraphics[width=0.8\linewidth,trim={0.4cm 14.3cm 0.3cm 0.3cm}, clip]{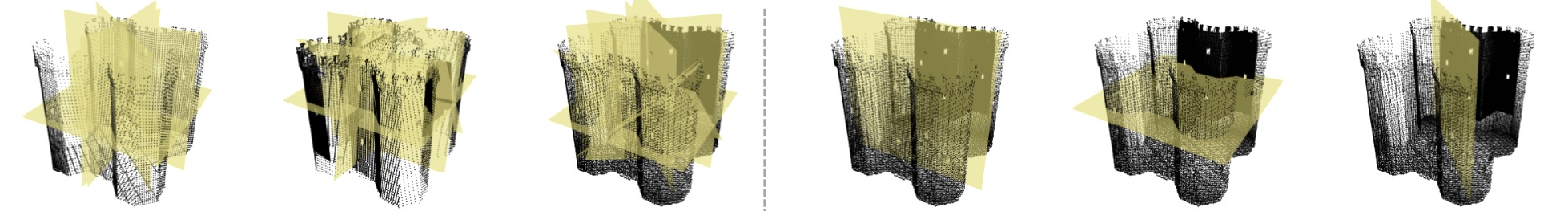}
    \caption{Robustness to sparse input. We show detection results on sparse-view reconstructed input point cloud, as well as the top 3 planes for the 8-view case. 
     }
    \label{fig:sparse}
    \vspace{-1em}
\end{figure*}

\end{document}